\documentclass[10pt,twocolumn]{article}

\usepackage{geometry}
\def\abstract
{
   \centerline{\large\bf Abstract}
   \vspace*{9pt}
   \it
}

\usepackage{times}
\usepackage{epsfig}
\usepackage{amsmath}
\usepackage{amssymb}
\usepackage{subfig}
\usepackage{amsthm}
\usepackage{amsfonts}
\usepackage{mathrsfs}
\usepackage{booktabs}
\usepackage{multirow, eucal}
\usepackage{helvet}
\usepackage{courier}

\usepackage[hyphens]{url}

\usepackage{graphicx}
\usepackage{cite}
\usepackage{subfig}
\usepackage{tabularx}

\usepackage[margin=4pt,font=footnotesize,labelfont=bf,labelsep=endash,tableposition=top]{caption}

\newcommand\T{\rule{0pt}{1.2ex}}
\newcommand\B{\rule[-0.8ex]{0pt}{0pt}}

\begin{document}
\title{
\Large \bf
NAS-FCOS: Fast Neural Architecture Search for Object Detection\thanks{
NW, YG, HC
 contributed to this work equally.}
}

\author{
Ning Wang$ ^1$, ~ Yang Gao$ ^1$, ~ Hao Chen$ ^2$, ~ Peng Wang$ ^1$, ~ Zhi Tian$ ^2$,
~ Chunhua Shen$ ^2$, ~ Yanning Zhang$ ^1$
\\[.152cm]
$ ^1 $School of Computer Science, Northwestern Polytechnical University, China\\
$ ^2 $School of Computer Science, The University of Adelaide, Australia
}

\date{}
\maketitle

\begin{abstract}
The success of deep neural networks relies on significant architecture engineering.
Recently neural architecture search (NAS) has emerged as a promise to greatly reduce manual effort in network design by automatically searching for optimal architectures, although typically such algorithms need an excessive amount of computational resources, e.g., a few thousand GPU-days. To date, on challenging vision tasks such as object detection, NAS, especially fast versions of NAS, is less studied. 
Here we propose to search for the decoder structure of object detectors with search efficiency being taken into consideration. To be more specific, we aim to efficiently search for the feature pyramid network (FPN) as well as the prediction head of a simple anchor-free object detector, namely FCOS~\cite{tian2019fcos}, using a tailored reinforcement learning paradigm.
With carefully designed search space, search algorithms and strategies for evaluating network quality, we are able to efficiently search a top-performing detection architecture within $4$ days using $8$ V100 GPUs.
The discovered architecture surpasses state-of-the-art object detection models (such as Faster R-CNN, RetinaNet and FCOS) by $1.5$ to $3.5$ points in AP on the COCO dataset,with comparable computation complexity and memory footprint, demonstrating the efficacy of the proposed NAS for object detection.

\end{abstract}

\section{Introduction}
Object detection is one of the fundamental tasks in computer vision, and has been researched extensively. 
In the past few years, state-of-the-art methods for this task are based on deep convolutional neural networks (such as Faster R-CNN \cite{ren2015faster}, RetinaNet~\cite{lin2017feature}), due to their impressive performance. Typically, the designs of object detection networks are much more complex than those for image classification, because the former need to localize and classify multiple objects in an image simultaneously while the latter only need to output image-level labels.
Due to its complex structure and numerous hyper-parameters, designing effective object detection networks is more challenging and usually needs much manual effort.

On the other hand, Neural Architecture Search (NAS) approaches~\cite{ghiasi2019fpn,nekrasov2018fast,zoph2016neural} have been showing impressive results on automatically discovering top-performing neural network architectures in large-scale search spaces.
Compared to manual designs, NAS methods are data-driven instead of experience-driven, and hence need much less human intervention.
As defined in~\cite{elsken2018neural}, the workflow of NAS can be divided into the following three processes:
$1$) sampling architecture from a search space following some search strategies;
$2$) evaluating the performance of the sampled architecture;
and
$3$) updating the parameters based on the performance.

One of the main problems prohibiting NAS from being used in more realistic applications is its search efficiency. The evaluation process is the most time consuming part because it involves a full training procedure of a neural network. To reduce the evaluation time, in practice a proxy task is often used as a lower cost substitution. In the proxy task, the input, network parameters and training iterations are often scaled down to speedup the evaluation. However, there is often a performance gap for samples between the proxy tasks and target tasks, which makes the evaluation process biased. How to design proxy tasks that are both accurate and efficient for specific problems is a challenging problem.
Another solution to improve search efficiency is constructing a supernet that covers the complete search space and training candidate architectures with shared parameters~\cite{liu2018darts, pham2018enas}.
However, this solution leads to significantly increased memory consumption and restricts itself to small-to-moderate sized search spaces.

To our knowledge, studies on efficient and accurate NAS approaches to object detection networks are rarely touched, despite its significant importance. 
To this end, we present a fast and memory saving NAS method for object detection networks, which is capable of 
discovering top-performing architectures within significantly reduced search time. 
Our overall detection architecture is based on FCOS~\cite{tian2019fcos},
a simple anchor-free one-stage object detection framework, in which the feature pyramid network and prediction head are searched using our proposed NAS method.

Our main contributions are summarized as follows.
\begin{itemize}
    \item In this work, we propose a fast and memory-efficient NAS method for searching both FPN and head architectures, 
    with carefully designed proxy tasks, search space and evaluation strategies, which is able to find top-performing architectures over $3,000$ architectures using $28$ GPU-days only.
    
    Specifically, this high efficiency is enabled with the following designs. 
    
    $-$ Developing a fast proxy task training scheme by skipping the backbone finetuning stage;
    
    $-$ Adapting progressive search strategy to reduce time cost taken by the extended search space;
    
    $-$ Using a more discriminative criterion for evaluation of searched architectures.
    
    $-$ Employing an efficient anchor-free one-stage detection framework with simple post processing;

    \item
    Using NAS, we explore the workload relationship between FPN and head, proving the importance of weight sharing in head.
    
    \item 
    We show that the overall structure of NAS-FCOS is general and flexible in that it can be equipped with various backbones including MobileNetV$2$, ResNet-$50$, ResNet-$101$ and ResNeXt-$101$, and surpasses state-of-the-art object detection algorithms using comparable computation complexity and memory footprint. More specifically, our model can improve the AP by $1.5\sim3.5$ points on all above models comparing to their FCOS counterparts.

\end{itemize}

\section{Related Work}

\subsection{Object Detection}
The frameworks of deep neural networks for object detection can be roughly categorized into two types:
one-stage detectors~\cite{lin2017focal} and two-stage detectors~\cite{he2017mask, ren2015faster}.

Two-stage detection frameworks first generate class-independent region proposals using a region proposal network (RPN),
and then classify and refine them using extra detection heads. In spite of achieving top performance, the two-stage methods have noticeable drawbacks: 
they are computationally expensive and have many hyper-parameters that need to be tuned to fit a specific dataset.

In comparison, the structures of one-stage detectors are much simpler.
They directly predict object categories and bounding boxes at each location of feature maps generated by a single CNN backbone.

Note that most state-of-the-art object detectors (including both one-stage detectors~\cite{lin2017focal,liu2016ssd, yolov3} and two-stage detectors~\cite{ren2015faster}) make predictions based on anchor boxes of different scales and aspect ratios at each convolutional feature map location. 
However, the usage of anchor boxes may lead to high imbalance between object and non-object examples and introduce extra hyper-parameters. 
More recently, anchor-free one-stage detectors~\cite{kong2019foveabox,law2018cornernet,tian2019fcos,zhou2019objects,zhu2019fsaf} have attracted increasing research interests, due to their simple fully convolutional architectures and reduced consumption of computational resources.

\subsection{Neural Architecture Search}
NAS is usually time consuming. We have seen great improvements from $24,000$ GPU-days~\cite{zoph2016neural} to $0.2$ GPU-day~\cite{zhou2019bayesnas}. The trick is to first construct a supernet containing the complete search space and train the candidates all at once with bi-level optimization and efficient weight sharing~\cite{liu2019auto,liu2018darts}. But the large memory allocation and difficulties in approximated optimization prohibit the search for more complex structures.

Recently researchers~\cite{cai2018proxylessnas,guo2019single,stamoulis2019single} propose to apply single-path training to reduce the bias introduced by approximation and model simplification of the supernet. DetNAS~\cite{chen2019detnas}
follows this idea to search for an efficient object detection architecture.
One limitation of the single-path approach is that the search space is restricted to a sequential structure. Single-path sampling and straight through estimate of the weight gradients introduce large variance to the optimization process and prohibit the search for more complex structures under this framework. Within this very simple search space, NAS algorithms can only make trivial decisions like kernel sizes for manually designed modules.

Object detection models are different from single-path image classification networks in their way of merging multi-level features and distributing the task to parallel prediction heads. Feature pyramid networks (FPNs)
~\cite{ghiasi2019fpn,Alexander2019panoptic,lin2017feature,Liu2019AnEnd,zhao2019pyramid}, designed to handle this job, plays an important role in modern object detection models. NAS-FPN~\cite{ghiasi2019fpn} targets on searching for an FPN alternative based on one-stage framework RetinaNet~\cite{lin2017focal}. Feature pyramid architectures are sampled with a recurrent neural network (RNN) controller. The RNN controller is trained with reinforcement learning (RL). However, the search is very time-consuming even though a proxy task with ResNet-10 backbone is trained to evaluate each architecture.

Since all these three kinds of research (~\cite{chen2019detnas,ghiasi2019fpn} and ours) focus on object detection framework, we demonstrate the differences among them that
{\it DetNAS~\cite{chen2019detnas} aims to search for the designs of better backbones, while NAS-FPN~\cite{ghiasi2019fpn} searches the FPN structure, and our search space contains both FPN and head structure.}

To speed up reward evaluation of RL-based NAS, the work of~\cite{nekrasov2018fast} proposes to use progressive tasks and other training acceleration methods. By caching the encoder features, they are able to train semantic segmentation decoders with very large batch sizes very efficiently. In the sequel of this paper, we refer to this technique as fast decoder adaptation. However, directly applying this technique to object detection tasks does not enjoy similar speed boost, because they are either not in using a fully-convolutional model~\cite{lin2017feature} or require complicated post processing that are not scalable with the batch size~\cite{lin2017focal}.

To reduce the post processing overhead, we resort to a recently introduced anchor-free one-stage framework, namely, FCOS~\cite{tian2019fcos}, which significantly improve the search efficiency by cancelling the processing time of anchor-box matching in RetinaNet.

Compared to its anchor-based counterpart, FCOS significantly reduces the training memory footprint while being able to improve the performance.

\section{Our Approach}

In our work, we search for anchor-free fully convolutional detection models with fast decoder adaptation. Thus, NAS methods can be easily applied.

\subsection{Problem Formulation}
We base our search algorithm upon a one-stage framework FCOS due to its simplicity. Our training tuples $\{(\mathbf x, Y)\}$ consist of input image tensors $\mathbf x$ of size $(3\times H\times W)$ and FCOS output targets $Y$ in a pyramid representation, which is a list of tensors $\mathbf y_l$ each of size $((K+4+1)\times H_l\times W_l)$ where $H_l\times W_l$ is feature map size on level $p$ of the pyramid. $(K+4+1)$ is the output channels of FCOS, the three terms are length-$K$ one-hot classification labels, $4$ bounding box regression targets and $1$ centerness factor respectively.

The network $g: \mathbf x\rightarrow \hat{Y}$ in original FCOS consists of three parts, a backbone $b$, FPN $f$ and multi-level subnets we call prediction heads $h$ in this paper. First backbone $b: \mathbf x\rightarrow C$ maps the input tensor to a set of intermediate-leveled features $C = \{\mathbf c_3, \mathbf c_4, \mathbf c_5\}$, with resolution $(H_i\times W_i) = (H/2^i \times W/2^i)$. Then FPN $f: C\rightarrow P$ maps the features to a feature pyramid $P=\{\mathbf p_3, \mathbf p_4, \mathbf p_5, \mathbf p_6, \mathbf p_7\}$. Then the prediction head $h: \mathbf p\rightarrow \mathbf y$ is applied to each level of $P$ and the result is collected to create the final prediction. To avoid overfitting, same $h$ is often applied to all instances in $P$.

Since objects of different scales require different effective receptive fields, the mechanism to select and merge intermediate-leveled features $C$ is particularly important in object detection network design. Thus, most researches~\cite{liu2016ssd,ren2015faster} are carried out on designing $f$ and $h$ while using widely-adopted backbone structures such as ResNet~\cite{he2016identity}. Following this principle, our search goal is to decide when to choose which features from $C$ and how to merge them.

To improve the efficiency, we reuse the parameters in $b$ pretrained on target dataset and search for the optimal structures after that. For the convenience of the following statement, we call the network components to search for, namely $f$ and $h$, together the decoder structure for the objection detection network.

$f$ and $h$ take care of different parts of the detection job. $f$ extracts features targeting different object scales in the pyramid representations $P$, while $h$ is a unified mapping applied to each feature in $P$ to avoid overfitting. In practice, people seldom discuss the possibility of using a more diversified $f$ to extract features at different levels or how many layers in $h$ need to be shared across the levels. In this work, we use NAS as an automatic method to test these possibilities.

\subsection{Search Space}
Considering the different functions of $f$ and $h$, we apply two search space respectively. Given the particularity of FPN structure, a basic block with new overall connection and $f$'s output design is built for it. For simplicity, sequential space is applied for $h$ part.

We replace the cell structure with atomic operations to provide even more flexibility. To construct one basic block, we first choose two layers $\mathbf x_1$, $\mathbf x_2$ from the sampling pool $X$ at \texttt{id1}, \texttt{id2}, then two operations \texttt{op1}, \texttt{op2} are applied to each of them and an aggregation operation \texttt{agg} merges the two output into one feature. To build a deep decoder structure, we apply multiple basic blocks with their outputs added to the sampling pool. Our basic block $bb_t: X_{t-1}\rightarrow X_t$ at time step $t$ transforms the sampling pool $X_{t-1}$ to $X_t = X_{t-1}\cup \{\mathbf{x}_t\}$, where $\mathbf{x}_t$ is the output of $bb_t$.

\begin{table}[t]
	\begin{center}
	    \begin{tabular}{c|c}
		    \hline
			ID & Description \T\B\\
			\hline
			0 & separable conv $3\times3$\T\B\\
			\hline
			1 & separable conv $3\times3$ with dilation rate $3$\T\B\\
			\hline
			2 & separable conv $5\times5$ with dilation rate $6$\T\B\\
			\hline
			3 & skip-connection\T\B\\
			\hline
			4 & deformable $3\times3$ convolution\B\\
            \hline
		\end{tabular}
		\caption{Unary operations used in the search process.
			\label{table:unary}}
	\end{center}
\end{table}

The candidate operations are listed in Table~\ref{table:unary}. We include only separable/depth-wise
convolutions so that the decoder can be efficient. In order to enable the decoder to apply convolutional filters on irregular grids, here we have also included deformable $3\times3$ convolutions~\cite{zhu2018deformable}.
For the aggregation operations, we include element-wise sum and
concatenation followed by a $1\times1$ convolution.

\begin{figure*}[thb!]
	\centering
	\subfloat{\includegraphics[width = 0.95\linewidth]{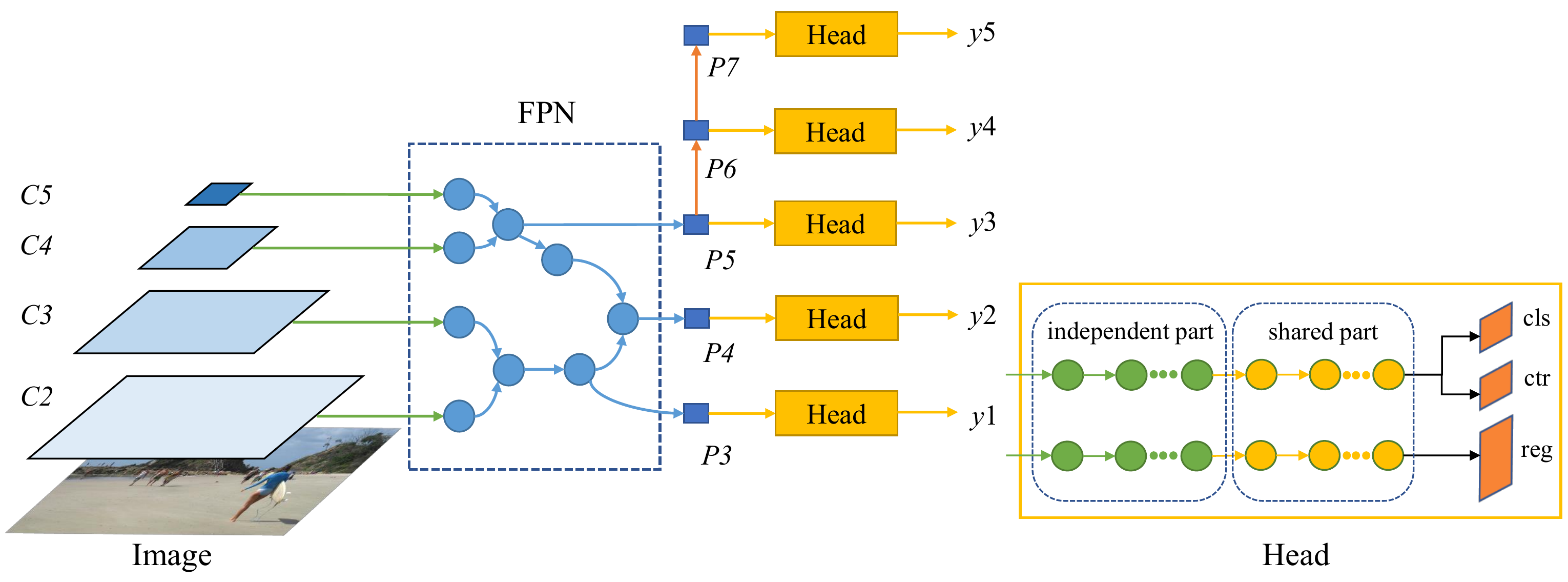}}
	\caption{A conceptual example of our NAS-FCOS decoder. It consists of two sub networks, an FPN $f$ and a set of prediction heads $h$ which have shared structures. One notable difference with other FPN-based one-stage detectors is that our heads have partially shared weights. Only the last several layers of the predictions heads (marked as yellow) are tied by their weights. The number of layers to share is decided automatically by the search algorithm. Note that both FPN and head are in our actual search space; and have more layers than shown in this figure. Here the figure is for illustration only.
\label{fig:main}}
\end{figure*}

The decoder configuration can be represented by a sequence with three components, FPN configuration, head configuration and weight sharing stages. We provide detailed descriptions to each of them in the following sections. The complete diagram of our decoder structure is shown in Fig.~\ref{fig:main}.

\subsubsection{FPN Search Space}
As mentioned above, the FPN $f$ maps the convolutional features $C$ to $P$. First, we initialize the sampling pool as $X_0 = C$. Our FPN is defined by applying the basic block $7$ times to the sampling pool, $f:= bb_1^f\circ bb_2^f \circ \cdots \circ bb_7^f$. To yield pyramid features $P$, we collect the last three basic block outputs $\{\mathbf x_5, \mathbf x_6, \mathbf x_7\}$ as $\{\mathbf p_3, \mathbf p_4, \mathbf p_5\}$.

To allow shared information across all layers, we use a simple rule to create global features. If there is some dangling layer $\mathbf x_t$ which is not sampled by later blocks $\{bb_i^f|i > t\}$ nor belongs to the last three layers $t < 5$, we use element-wise add to merge it to all output features
\begin{align}
    \mathbf{p}^*_i = \mathbf p_i + \mathbf x_t, \,\, i\in\{3, 4, 5\}.
\end{align}
Same as the aggregation operations, if the features have different resolution, the smaller one is upsampled with bilinear interpolation.

To be consistent with FCOS, $\mathbf{p}_6$ and $\mathbf{p}_7$ are obtained via a $3\times3$ stride-$2$ convolution on $\mathbf{p}_5$ and $\mathbf{p}_6$ respectively.

\subsubsection{Prediction Head Search Space}

Prediction head $h$ maps each feature in the pyramid $P$ to the output of corresponding $\mathbf y$, which in FCOS and RetinaNet, consists of four $3\times3$ convolutions. To explore the potential of the head, we therefore extend a sequential search space for its generation. Specifically, our head is defined as a sequence of six basic operations. Compared with candidate operations in the FPN structures, the head search space has two slight differences. First, 
we add standard convolution modules (including conv$1$x$1$ and conv$3$x$3$) to the head sampling pool for better comparison. Second, we follow the design of FCOS by replacing all the Batch Normalization (BN) layers to Group Normalization (GN)~\cite{wu2018group} in the operations sampling pool of head, considering that head needs to share weights between different levels, which causes BN invalid. The final output of head is the output of the last (sixth) layer.

\subsubsection{Searching for Head Weight Sharing}

To add even more flexibility and understand the effect of weight sharing in prediction heads, 
we further add an index $i$
as the location where the prediction head starts to share weights. For every layer before stage $i$, the head $h$ will create independent set of weights for each FPN output level, otherwise, it will use the global weights for sharing purpose.         

Considering the independent part of the heads being extended FPN branch and the shared part as head with adaptive-length, we can further balance the workload for each individual FPN branch to extract level-specific features and the prediction head shared across all levels.

\subsection{Search Strategy}
\label{search strategy}
RL based strategy is applied to the search process. We rely on an LSTM-based controller to predict the full configuration.
We consider using a progressive search strategy rather than the joint search for both FPN structure and prediction head part, since the former requires less computing resources and time cost than the latter. The training dataset is randomly split into a meta-train $D_t$ and meta-val $D_v$ subset. To speed up the training, we fix the backbone network and cache the pre-computed backbone output $C$. This makes our single architecture training cost independent from the depth of backbone network. Taking this advantage, we can apply much more complex backbone structures and utilize high quality multilevel features as our decoder's input. We find that the process of backbone finetuning can be skipped if the cached features are powerful enough. Speedup techniques such as Polyak weight averaging are also applied during the training.

The most widely used detection metric is average precision (AP). However, due to the difficulty of object detection task, at the early stages, AP is too low to tell the good architectures from the bad ones, which makes the controller take much more time to converge. To make the architecture evaluation process easier even at the early stages of the training, we therefore use negative loss sum as the reward instead of average precision:
\begin{equation}
\begin{split}
    R(a) = &- \sum_{(x, Y)\in D_v}(L_{cls}(x, Y|a) \\
          &+ L_{reg}(x, Y|a) + L_{ctr}(x, Y|a))
\end{split}
\label{reward}
\end{equation}
where $L_{cls}$, $L_{reg}$, $L_{ctr}$ are the three loss terms in FCOS. Gradient of the controller is estimated via proximal policy optimization (PPO)~\cite{schulman2017proximal}.
\label{search strategy}

\section{Experiments}

\subsection{Implementation Details}

\subsubsection{Searching Phase}

We design a fast proxy task for evaluating the decoder architectures sampled in the searching phase.
PASCAL VOC is selected as the proxy dataset, which contains $5715$ training images with object bounding box annotations of $20$ classes. Transfer capacity of the structures can be illustrated since the search and full training phase use different datasets.
The VOC training set is randomly split into a meta-train set with $4,000$ images and a meta-val set with $1715$ images.
For each sampled architecture, we train it on meta-train and compute the reward~\eqref{reward} on meta-val.
Input images are resized to short size $384$ and then randomly cropped to $384\times384$. Target object sizes of interest are scaled correspondingly. We use Adam optimizer with learning rate $8$e$-4$ and batch size $200$. Polyak averaging is applied with the decay rates of $0.9$. The decoder is evaluated after $300$ iterations.
As we use fast decoder adaptation, the backbone features are fixed and cached during the search phase.
To enhance the cached backbone features, we first initialize them with pre-trained weights provided by open-source implementation of FCOS\footnote{https://tinyurl.com/FCOSv1} and then finetune on VOC using the training strategies of FCOS. Note that the above finetuning process is only performed once at the begining of the search phase.  

A progressive strategy is used for the search of $f$ and $h$. We first search for the FPN part and retain the original head. All operations in the FPN structure have $64$ output channels. The decoder inputs $C$ are resized to fit output channel width of FPN via $1\times1$ convolutions. After this step, a searched FPN structure is fixed and the second stage searching for the head will be started based on it. Most parameters for searching head are identical to those for searching FPN structure, with the exception that the output channel width is adjusted from $64$ to $128$ to deliver more information.

\begin{figure}[t!]
\centering
\includegraphics[width=0.50\textwidth]{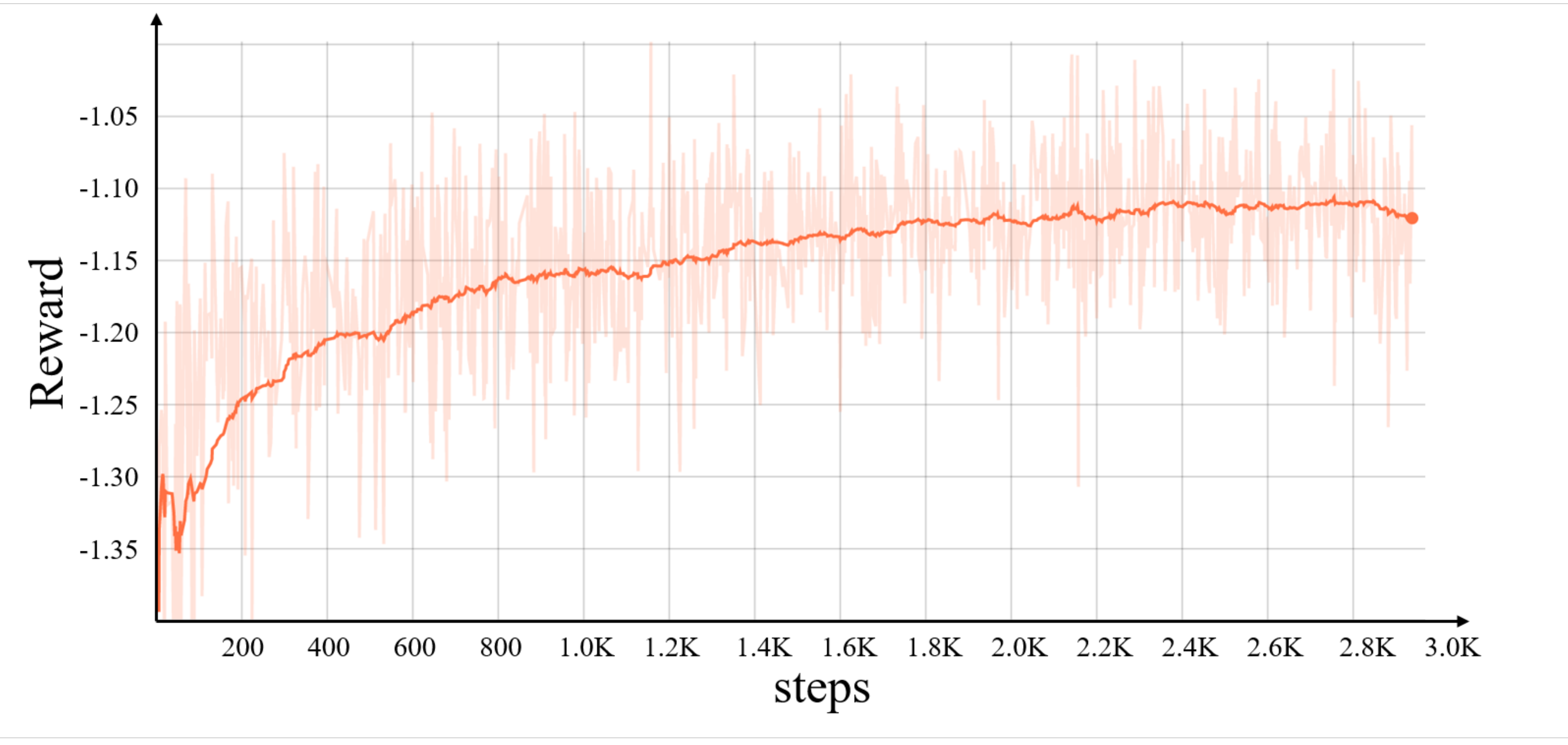}
\caption{Performance of reward during the proxy task, which has been growing throughout the process, indicating that the model of reinforcement learning works.}
\label{fig:reward}
\end{figure}

For the FPN search part, the controller model nearly converged after searching over $2.8$K architectures on the proxy task as shown in Fig.~\ref{fig:reward}. Then, the top-$20$ best performing architectures on the proxy task are selected for the next full training phase.
For the head search part, we choose the best searched FPN among the top-$20$ architectures and pre-fetch its features. It takes about $600$ rounds for the controller to nearly converge, which is much faster than that for searching FPN architectures. After that, we select for full training the top-$10$ heads that achieve best performance on the proxy task. In total, the whole search phase can be finished within $4$ days using $8$ V100 GPUs.

\begin{table*}[th!]
\centering
\scalebox{0.95}{
    \begin{tabular}{ll|cc|cc}
    \hline\noalign{\smallskip}
    Decoder & Backbone & FLOPs (G) & Params (M) & AP\\
    \noalign{\smallskip}\hline\noalign{\smallskip}
    FPN-RetinaNet @$256$ & MobileNetV$2$ & $133.4$ & $11.3$ & $30.8$\\
    FPN-FCOS @$256$ & MobileNetV$2$ & $105.4$ & $9.8$ & $31.2$\\
    NAS-FCOS (ours) @$128$ & MobileNetV$2$ & $\mathbf{39.3}$ & $\mathbf{5.9}$ & $32.0$\\
    NAS-FCOS (ours) @$128$-$256$ & MobileNetV$2$ & $95.6$ & $9.9$ & $33.8$\\
    NAS-FCOS (ours) @$256$ & MobileNetV$2$ & $121.8$ & $16.1$ & $\mathbf{34.7}$\\
    \noalign{\smallskip}\hline\noalign{\smallskip}
    FPN-RetinaNet @$256$ & R-$50$ & $198.0$ & $33.6$ & $36.1$\\
    FPN-FCOS @$256$ & R-$50$ & $169.9$ & $32.0$ & $37.4$\\

    NAS-FCOS (ours) @$128$ & R-$50$ & $\mathbf{104.0}$ & $\mathbf{27.8}$ & $37.9$\\
    NAS-FCOS (ours) @$128$-$256$ & R-$50$ & $160.4$ & $31.8$ & $39.1$\\
    NAS-FCOS (ours) @$256$ & R-$50$ & $189.6$ & $38.4$ & $\mathbf{39.8}$\\

    \noalign{\smallskip}\hline\noalign{\smallskip}
    FPN-RetinaNet @$256$ & R-$101$ & $262.4$ & $52.5$ & $37.8$\\
    FPN-FCOS @$256$ & R-$101$ & $\mathbf{234.3}$ & $\mathbf{50.9}$ & $41.5$\\

    NAS-FCOS (ours) @$256$ & R-$101$ & $254.0$ & $57.3$ & $\mathbf{43.0}$\\
    \noalign{\smallskip}\hline\noalign{\smallskip}

    FPN-FCOS @$256$ & X-$64$x$4$d-$101$ & $371.2$ & $89.6$ & $43.2$ \\
    NAS-FCOS (ours) @$128$-$256$ & X-$64$x$4$d-$101$ & $\mathbf{361.6}$ & $\mathbf{89.4}$ & $\mathbf{44.5}$\\
    \noalign{\smallskip}\hline\noalign{\smallskip}
    FPN-FCOS @$256$ w/improvements & X-$64$x$4$d-$101$ & $371.2$ & $89.6$ & $44.7$ \\    
    NAS-FCOS (ours) @$128$-$256$ w/improvements & X-$64$x$4$d-$101$ & $\mathbf{361.6}$ & $\mathbf{89.4}$ & $\mathbf{46.1}$\\
    \noalign{\smallskip}\hline
    \end{tabular}
}
    \caption{Results on test-dev set of MS COCO after full training. R-$50$ and R-$101$ represents ResNet backbones and X-$64$x$4$d-$101$ represents ResNeXt-$101$ ($64\times4$d). All networks share the same input image resolution. FLOPs and parameters are being measured on $1088\times800$, which is the median of the input size on COCO. For RetinaNet and FCOS, we use official models provided by the authors. For our NAS-FCOS, @$128$ and @$256$ means that the decoder channel width is $128$ and $256$ respectively. @$128$-$256$ is the decoder with $128$ FPN width and $256$ head width. The same improving tricks used on the newest FCOS version are used in our model for fair comparison.
	\label{table:det}}
\end{table*}

\subsubsection{Full Training Phase}
In this phase, we fully train the searched models on the MS COCO training dataset, and select the best one by evaluating them on MS COCO validation images.  
Note that our training configurations are exactly the same as those in FCOS for fair comparison. Input images are resized to short size $800$ and the maximum long side is set to be $1333$. The models are trained using $4$ V100 GPUs with batch size $16$ for $90$K iterations. The initial learning rate is $0.01$ and reduces to one tenth at the $60$K-th and $80$K-th iterations. The improving tricks are applied only on the final model (w/improv).

\begin{figure}[h!]
\centering

\includegraphics[width=0.50\textwidth]{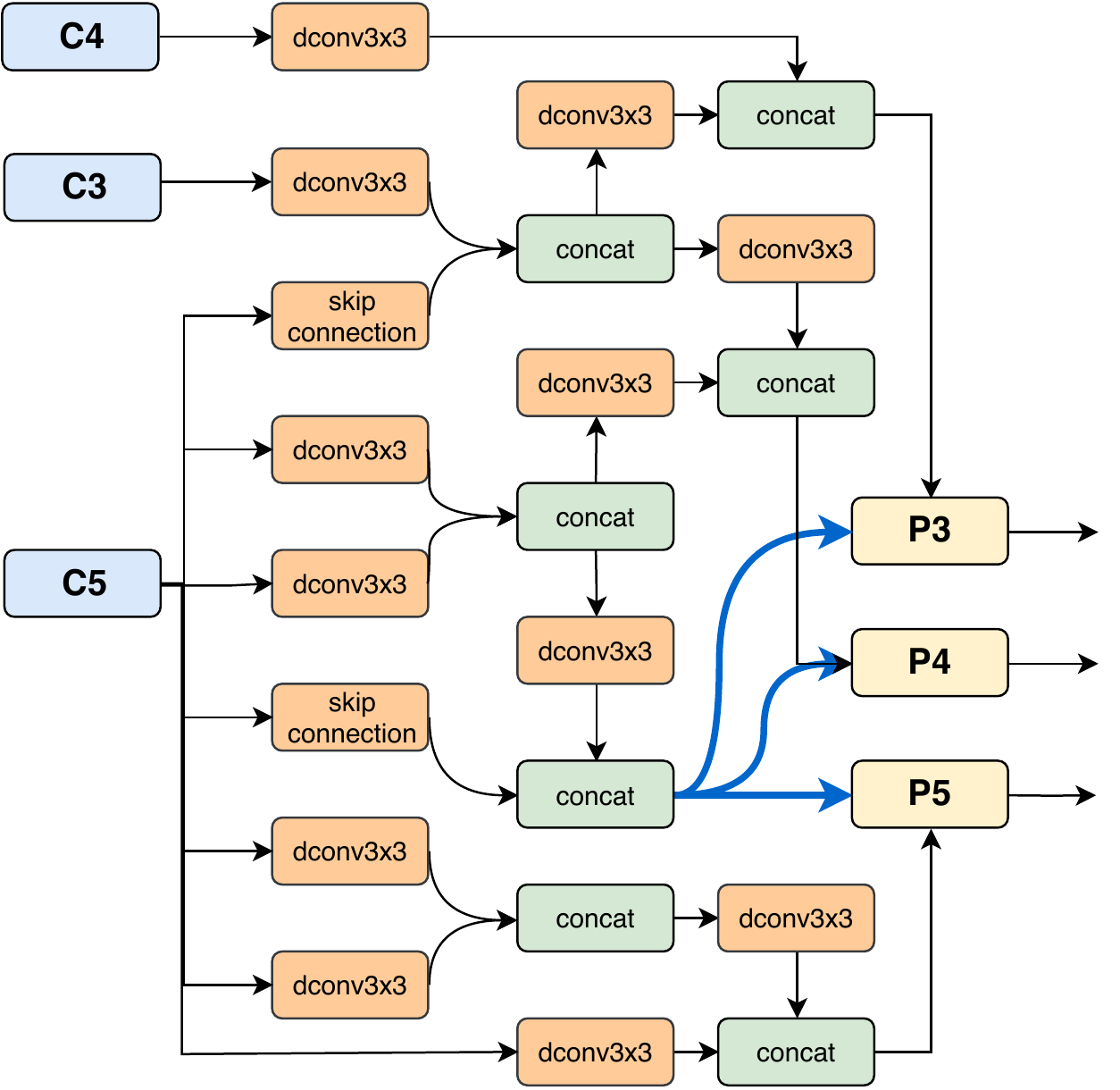}

\caption{Our discovered FPN structure. $C_2$ is omitted from this figure since it is not chosen by this particular structure during the search process.}
\label{fig:fpn}
\end{figure}

\subsection{Search Results}

\begin{figure}[t!]
\centering

\includegraphics[width=0.50\textwidth]{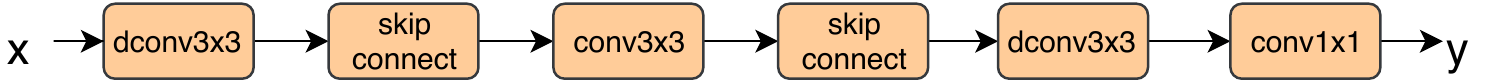}

\caption{Our discovered Head structure. 
}
\label{fig:searched_head}
\end{figure}

The best FPN structure is illustrated in Fig.~\ref{fig:fpn}. The controller identifies that deformable convolution and concatenation are the best performing operations for unary and aggregation respectively.
From Fig.~\ref{fig:searched_head}, we can see that the controller chooses to use $4$ operations (with two skip connections), rather than the maximum allowed $6$ operations. Note that the discovered ``dconv + $1$x$1$ conv'' structure achieves a good trade-off between accuracy and FLOPs.  
Compared with the original head, our searched head has fewer FLOPs/Params 
(FLOPs $79.24$G vs.\ $89.16$G, Params $3.41$M vs.\ $4.92$M) 
and significantly better performance (AP $38.7$ vs.\ $37.4$).

\begin{figure}[t!]
\centering
\includegraphics[width=0.48\textwidth]{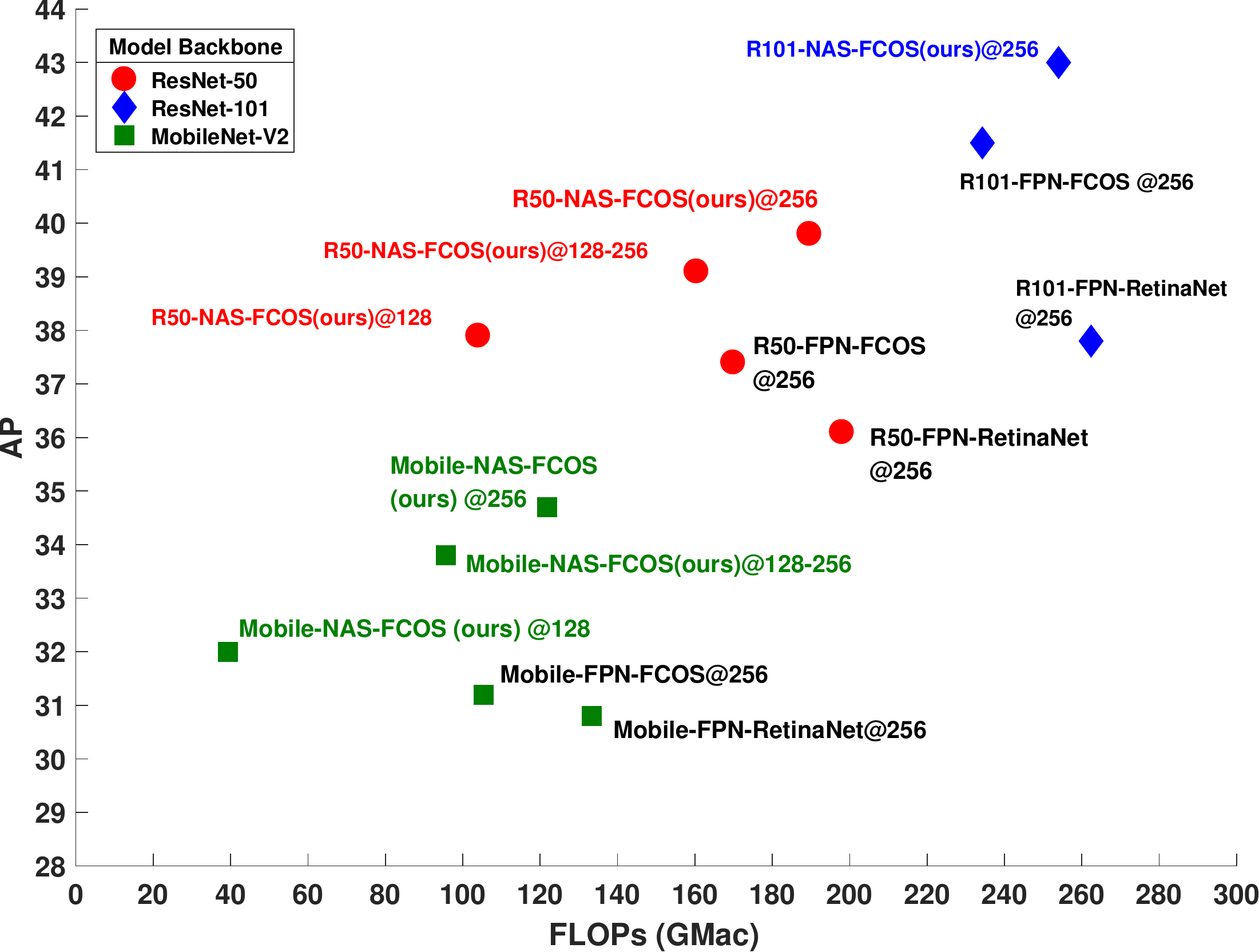}
\caption{Diagram of the relationship between FLOPs and AP with different backbones. Points of different shapes represent different backbones. NAS-FCOS@$128$ has a slight increase in precision which also gains the advantage of computation quantity. One with $256$ channels obtains the highest precision with more computation complexity. Using FPN channel width $128$ and prediction head $256$ (@$128$-$256$) offers a trade-off.
}
\label{fig:flops}
\end{figure}

\begin{figure}[t!]
\centering
\includegraphics[width=0.48\textwidth]{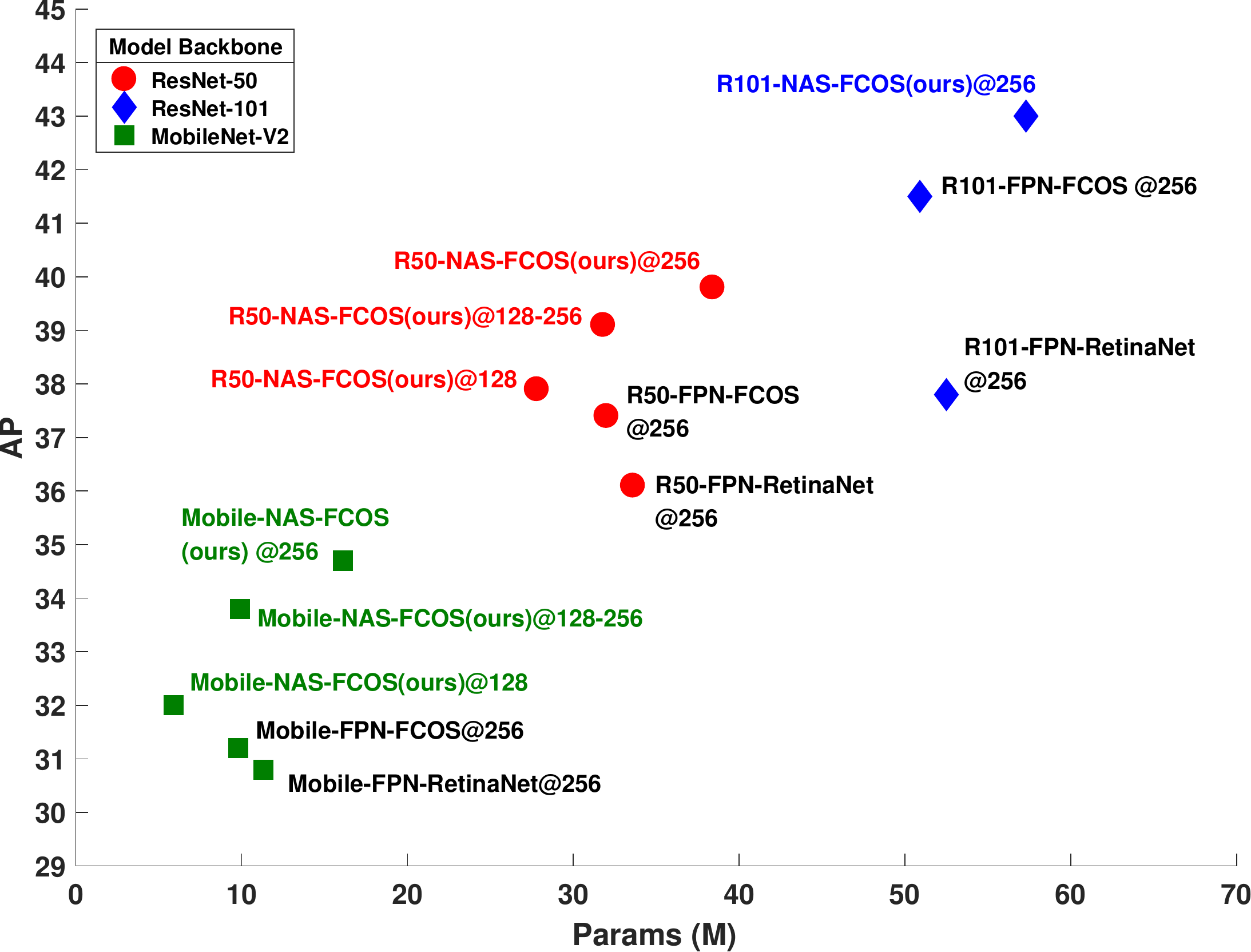}
\caption{Diagram of the relationship between parameters and AP with different backbones. Adjusting the number of channels in the FPN structure and head helps to achieve a balance between accuracy and parameters.}
\label{fig:params}
\end{figure}

We use the searched decoder together with either light-weight backbones 
such as MobileNet-V2~\cite{sandler2018mobilenetv2} 
or more powerful backbones such as ResNet-$101$~\cite{he2016identity} and ResNeXt-$101$~\cite{xie2016aggregated}. 
To balance the performance and efficiency, we implement three decoders with different computation budgets: one with feature dimension of $128$ (@$128$), one with $256$ (@$256$) and another with FPN channel width $128$ and prediction head $256$ (@$128$-$256$). The results on the COCO test-dev with short side being $800$ is shown in Table~\ref{table:det}. The searched decoder with feature dimension of $256$ (@$256$) surpasses its FCOS counterpart by $1.5$ to $3.5$ points in AP under different backbones. The one with $128$ channels (@$128$) has significantly reduced parameters and calculation, making it more suitable for resource-constrained environments. In particular, our searched model with $128$ channels and MobileNetV2 backbone suparsses the original FCOS with the same backbone by $0.8$ AP points with only $1/3$ FLOPS.
The third type of decoder (@$128$-$256$) achieves a good balance between accuracy and parameters. 
Note that our searched model outperforms the strongest FCOS variant by $1.4$ AP points ($46.1$ vs.\  $44.7$) with slightly smaller FLOPs and Params. The comparison of FLOPs and number of parameters with other models are illustrated in Fig.~\ref{fig:flops} and Fig.~\ref{fig:params} respectively.

\begin{figure}[t!]
\centering
\includegraphics[width=0.48\textwidth]{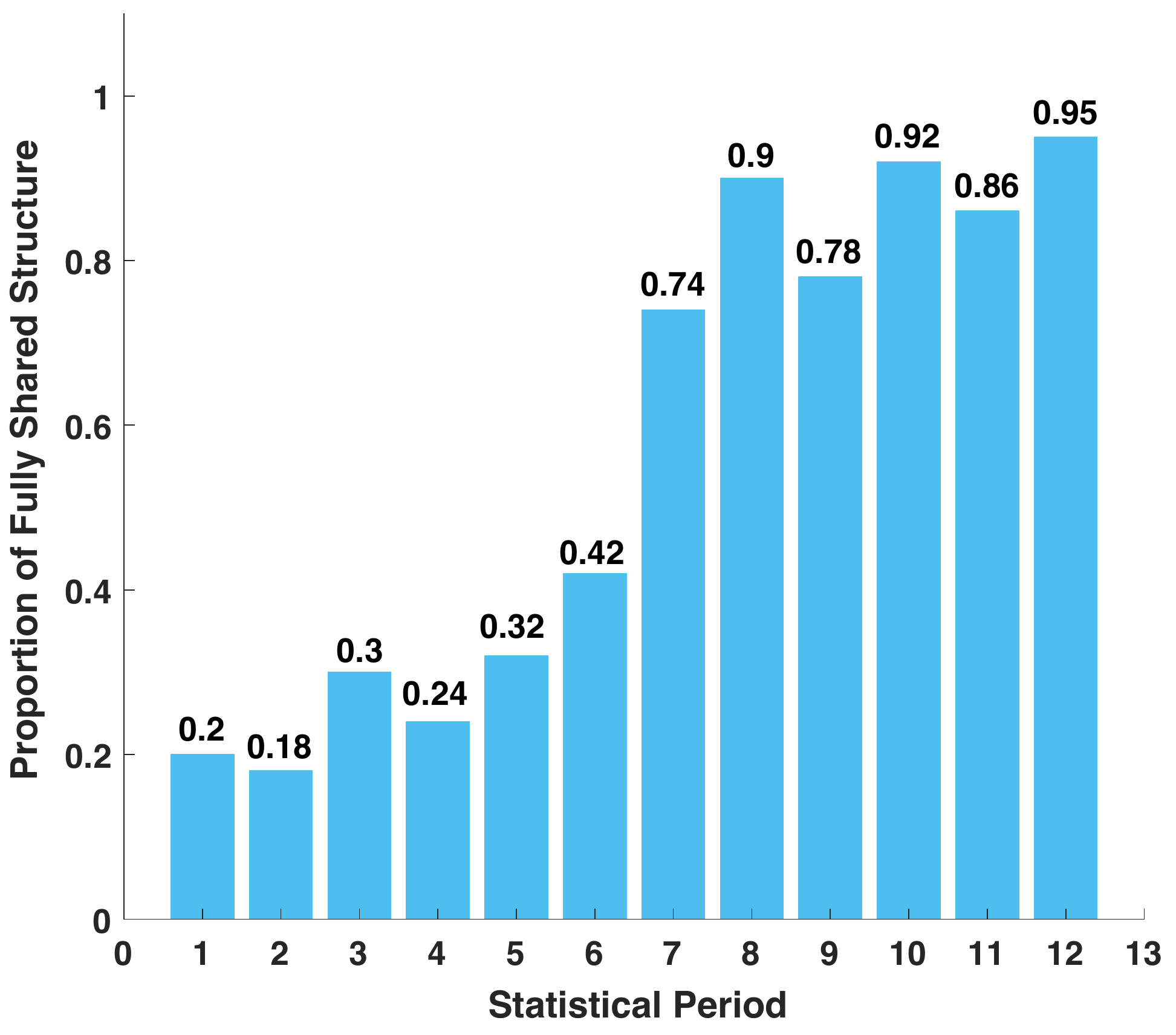}
\caption{Trend graph of head weight sharing during search. The coordinates in the horizontal axis represent the number of the statistical period. A period consists of $50$ head structures. The vertical axis represents the proportion of heads that fully share weights in $50$ structures.}
\label{fig:share_weights}
\end{figure}

In order to understand the importance of weight sharing in head, we add the number of layers shared by weights as an object of the search. Fig.~\ref{fig:share_weights} shows a trend graph of head weight sharing during search. We set $50$ structures as a statistical cycle. 
As the search deepens, the proportion of fully shared structures increases, indicating that on the multi-scale detection model, head weight sharing is a necessity.

\begin{table*}[t!]
\centering
\scalebox{0.95}{
    \begin{tabular}{l|c|c|c|c}
    \hline\noalign{\smallskip}
    Arch & FLOPs (G) & Search Cost (GPU-day) & Searched Archs & AP\\
    \noalign{\smallskip}\hline\noalign{\smallskip}
    NAS-FPN @$256$ R-$50$ & \textgreater$325.0$ & $333\times$\#TPUs & $17000$ & \textless$38.0$\\
    NAS-FPN $7$@$256$ R-$50$ & $1125.5$ & $333\times$\#TPUs & $17000$ & $44.8$\\
    DetNAS-FPN-Faster & - & $44$ & $2200$ & $40.2$\\
    DetNAS-RetinaNet & - & $44$ & $2200$ & $33.3$\\
    \noalign{\smallskip}\hline\noalign{\smallskip}
    NAS-FCOS (ours) @$256$ R-$50$ & $\mathbf{189.6}$ & $\mathbf{28}$ & $3000$ & $39.8$\\

    NAS-FCOS (ours) @$128$-$256$ X-$64$x$4$d-$101$  & $361.6$ & $\mathbf{28}$ & $3000$ & $\mathbf{46.1}$\\

    \noalign{\smallskip}\hline
    \end{tabular}
}
    \caption{ Comparison with other NAS methods.
    For NAS-FPN,
    the input size is $1280\times1280$ and
    the search cost should be timed by their number of TPUs used to train each architecture. 
    Note that 
    the FLOPs and AP of NAS-FPN @$256$ here are from Figure $11$ in NAS-FPN~\cite{ghiasi2019fpn},
    and NAS-FPN $7$@$256$ stacks the searched FPN structure $7$ times. 
    The input images are resized such that their shorter size is 800 pixels in DetNASNet~\cite{chen2019detnas} and our models. 
	\label{table:nas}}
\end{table*}

\begin{figure}[b!]
\centering
\includegraphics[width=0.48\textwidth]{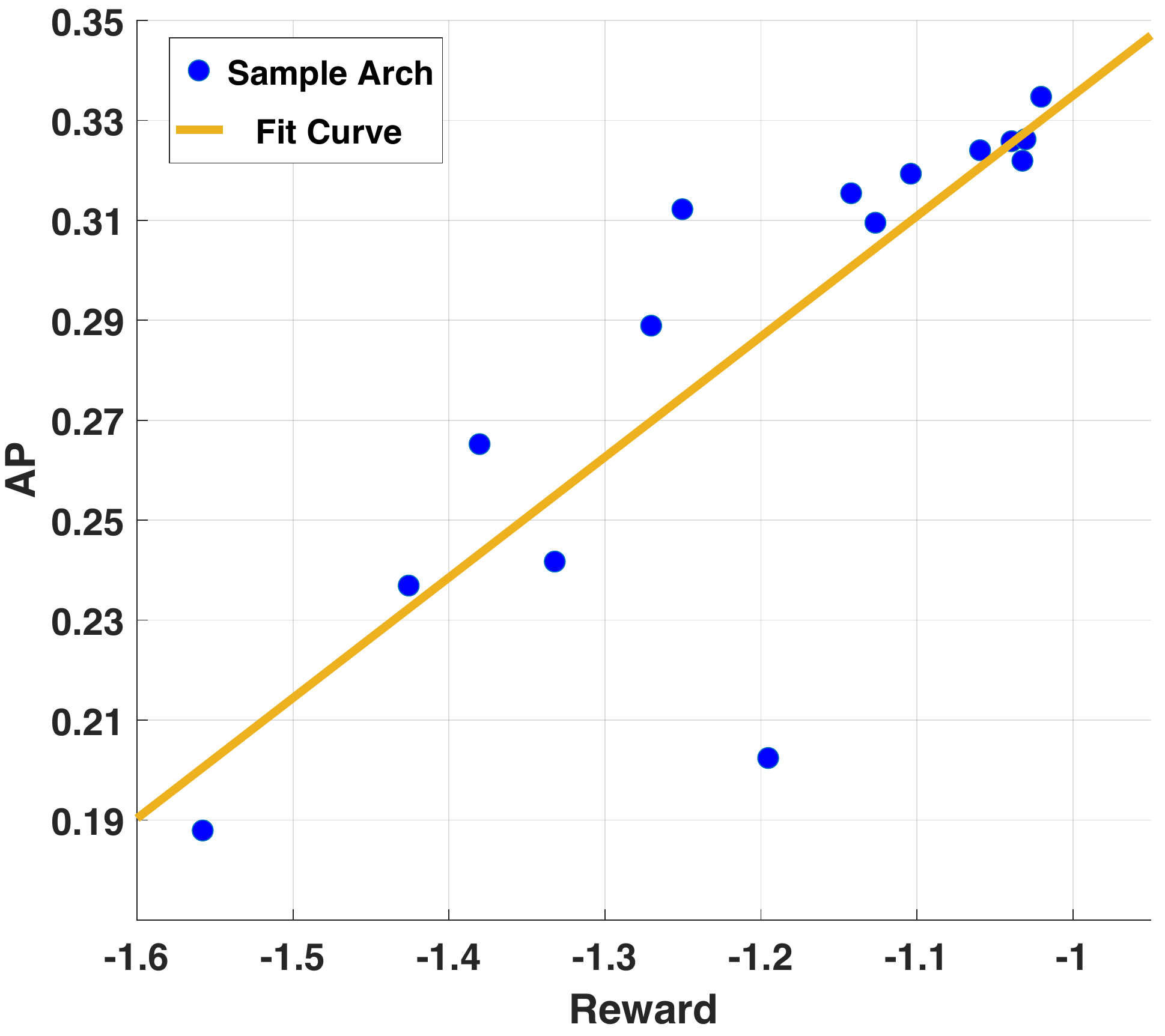}
\caption{Correlation between the search reward obtained on the VOC meta-val dataset and the AP evaluated on COCO-val.}
\label{fig:correlation}
\end{figure}

We also demonstrate the comparison with other NAS methods for object detection in Table~\ref{table:nas}. Our method is able to search for twice more architectures than DetNAS~\cite{chen2019detnas} per GPU-day. Note that the AP of NAS-FPN~\cite{ghiasi2019fpn} is achieved by stacking the searched FPN $7$ times, while we do not stack our searched FPN.
Our model with ResNeXt-101 ($64$x$4$d) as backbone outperforms NAS-FPN by $1.3$ AP points while using only $1/3$ FLOPs and less calculation cost.

We further measure the correlation between rewards obtained during the search process with the proxy dataset and APs attained by same architectures trained on COCO.
Specifically, we randomly sample $15$ architectures from all the searched structures trained on COCO with batch size $16$. Since full training on COCO is time-consuming, we reduce the iterations to $60$K. The model is then evaluated on the COCO $2017$ validation set. As visible in Fig.~\ref{fig:correlation}, there is a strong correlation between search rewards and APs obtained from COCO. Poor- and well-performing architectures can be distinguished by the  rewards on the proxy task very well.

\begin{figure}[t!]
\centering
\includegraphics[width=0.48\textwidth]{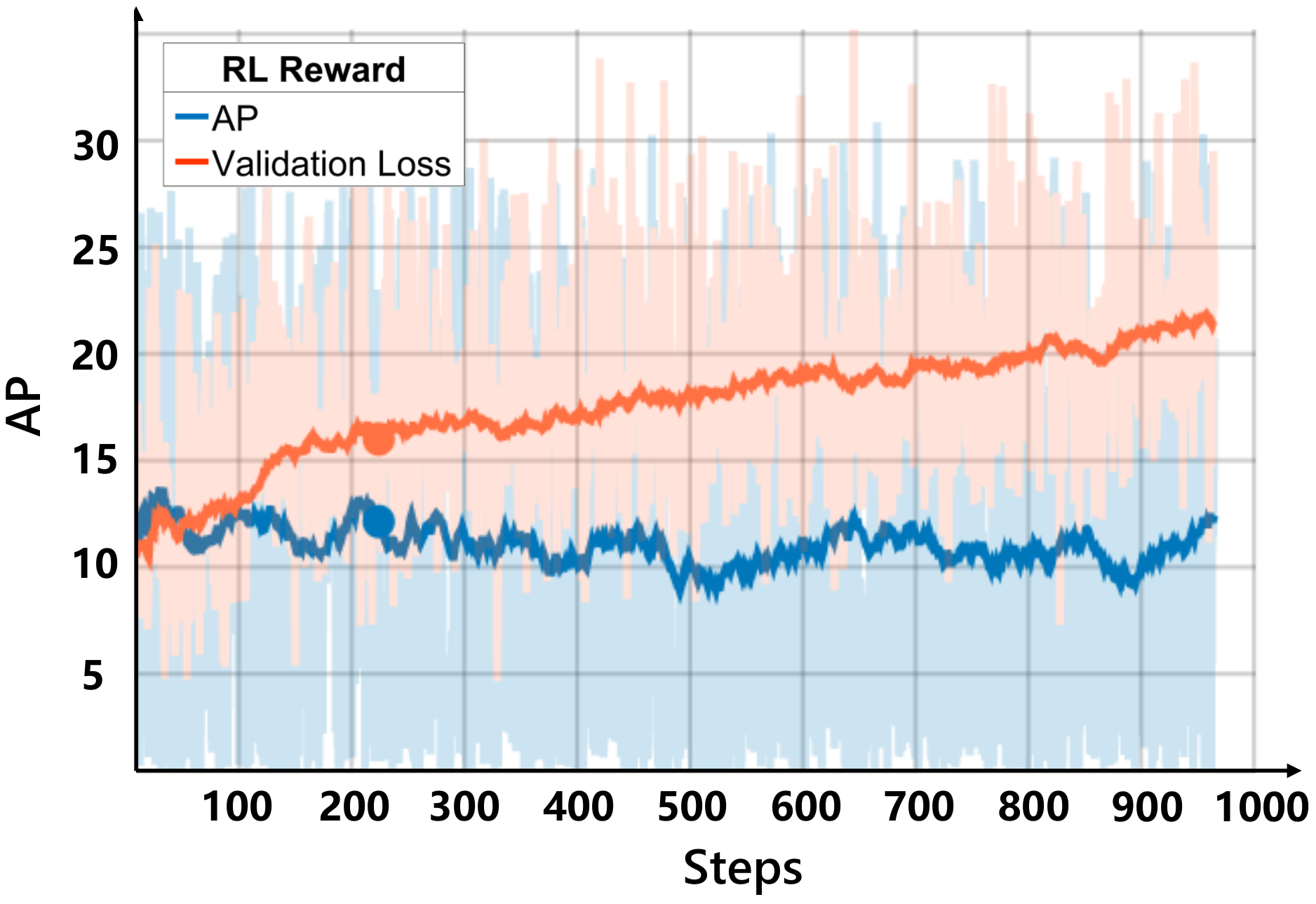}
\caption{Comparison of two different RL reward designs. The vertical axis represents AP obtained from the proxy task on the validation dataset.}
\label{fig:metric}
\end{figure}

\subsection{Ablation Study}

\subsubsection{Design of Reinforcement Learning Reward}

As we discussed above, it is common to use widely accepted indicators as rewards for specific tasks in the search, such as mIOU for segmentation and AP for object detection. However, we found that using AP as reward did not show a clear upward trend in short-term search rounds (blue curve in Fig.~\ref{fig:metric}). We further analyze the possible reason to be that the controller tries to learn a mapping from the decoder to the reward while the calculation of AP itself is complicated, which makes it difficult to learn this mapping within a limited number of iterations. In comparison, we clearly see the increase of AP with the validation loss as RL rewards (red curve in Fig.~\ref{fig:metric}). 

\begin{table}[t!]
\centering
\scalebox{0.95}{
    \begin{tabular}{c|c|c}
    \hline\noalign{\smallskip}
    Decoder & Search Space & AP\\
    \noalign{\smallskip}\hline\noalign{\smallskip}
    FPN-FCOS @$256$ & - & $37.4$\\
    \noalign{\smallskip}\hline\noalign{\smallskip}
    NAS-FCOS @$256$ & $h$ only & $38.7$\\
    NAS-FCOS @$256$ & $f$ only & $38.9$\\
    NAS-FCOS @$256$ & $f$ + $h$ & $\mathbf{39.8}$ \\
    \noalign{\smallskip}\hline\noalign{\smallskip}
    \end{tabular}
}
    \caption{Comparisons between APs obtained under different search space with ResNet-50 backbone.
	\label{table:effective}}
\end{table}

\subsubsection{Effectiveness of Search Space}
To further discuss the impact of the search spaces $f$ and $h$, we design three experiments for verification. One is to search $f$ with the original head being fixed, one is to search $h$ with the original FPN being fixed and another is to search the entire decoder ($f$+$h$). As shown in Table~\ref{table:effective}, it turns out that searching $f$ brings slightly more benefits than searching $h$ only. And our progressive search which combines both $f$ and $h$ achieves a better result.

\subsubsection{Impact of Deformable Convolution}
As aforementioned, deformable convolutions are included in the set of candidate operations for both $f$ and $h$, which are able to adapt to the geometric variations of objects. For fair comparison, we also replace the whole standard $3\times3$ convolutions with deformable $3\times3$ convolutions in FPN structure of the original FCOS and repeat them twice, making the FLOPs and parameters nearly equal to our searched model. 
The new model is therefore called DeformFPN-FCOS. 
It turns out that our NAS-FCOS model still achieves better performance (AP $= 38.9$ with FPN search only, and AP $= 39.8$ with both FPN and Head searched) than the DeformFPN-FCOS model (AP $= 38.4$) under this circumstance.

\section{Conclusion}
In this paper, we have proposed to use Neural Architecture Search to further optimize the process of designing object detection networks.
It is shown in this work that top-performing detectors can be efficiently searched using carefully designed proxy tasks, search strategies and model evaluation metrics.
The experiments on COCO demonstrates the efficiency of our discovered model NAS-FCOS and its flexibility to be used with various backbone architectures.

{\small
\bibliographystyle{reference}
\bibliography{draft}
}

\end{document}